\documentclass[sigconf]{acmart}
\acmSubmissionID{cfp6597}
\AtBeginDocument{%
  }

\copyrightyear{2025}
\acmYear{2025}
\setcopyright{cc}
\setcctype{by}
\acmConference[CIKM '25]{Proceedings of the 34th ACM International Conference on Information and Knowledge Management}{November 10--14, 2025}{Seoul, Republic of Korea}
\acmBooktitle{Proceedings of the 34th ACM International Conference on Information and Knowledge Management (CIKM '25), November 10--14, 2025, Seoul, Republic of Korea}\acmDOI{10.1145/3746252.3761272}
\acmISBN{979-8-4007-2040-6/2025/11}

\usepackage{multirow}
\usepackage{makecell}
\usepackage{xcolor}
\usepackage{colortbl}
\usepackage{booktabs}
\usepackage{algorithm}
\usepackage{algpseudocode}
\usepackage{amsmath}
\usepackage{defs}
\usepackage{enumitem}

\newcommand{\sysname}{\texttt{TFMAdapter}}
\begin{document}

\newcommand{\ctitle}{TFMAdapter: Lightweight Instance-Level Adaptation of Foundation Models for Forecasting with Covariates}
\title[\ctitle]{\ctitle}


\author{Afrin Dange}
\orcid{0009-0002-9532-3664}
\affiliation{%
  \institution{Indian Institute of Technology Bombay}
  \department{Centre for Machine Intelligence and Data Science}
  \city{Mumbai}
  \state{MH}
  \country{India}}
\email{dangeafrin@iitb.ac.in}

\author{Sunita Sarawagi}
\orcid{0009-0005-9538-6616}
\affiliation{%
  \institution{Indian Institute of Technology Bombay}
  \department{Department of Computer Science and Engineering}
  \city{Mumbai}
  \state{MH}
  \country{India}}
\email{sunita@iitb.ac.in}







\begin{abstract}
Time Series Foundation Models (TSFMs) have recently achieved state-of-the-art performance in univariate forecasting on new time series simply by conditioned on a brief history of past values. Their success demonstrates that large-scale pretraining across diverse domains can acquire the inductive bias to generalize from temporal patterns in a brief history. However, most TSFMs are unable to leverage covariates—future-available exogenous variables critical for accurate forecasting in many applications—due to their domain-specific nature and the lack of associated inductive bias.

We propose \sysname{}, a lightweight, instance-level adapter that augments TSFMs with covariate information without fine-tuning. Instead of retraining, \sysname{} operates on the limited history provided during a single model call, learning a non-parametric cascade that combines covariates with univariate TSFM forecasts.  However, such learning would require univariate forecasts at all steps in the history, requiring too many calls to the TSFM. To enable training on the full historical context while limiting TSFM invocations, \sysname{} uses a two-stage method: (1) generating pseudo-forecasts with a simple regression model, and (2) training a Gaussian Process regressor to refine predictions using both pseudo- and TSFM forecasts alongside covariates.

Extensive experiments on real-world datasets demonstrate that \sysname{} consistently outperforms both foundation models and supervised baselines, achieving a 24--27\% improvement over base foundation models with minimal data and computational overhead. Our results highlight the potential of lightweight adapters to bridge the gap between generic foundation models and domain-specific forecasting needs.

\end{abstract}

\ccsdesc[500]{Computing methodologies~Machine learning algorithms}

\keywords{Time Series Forecasting, Covariates, Foundation Models, Adaptation, Regression, Gaussian Processes}


  \maketitle

\section{Introduction}
\footnotetext[0]{Code is available at \href{https://github.com/AfrinDange/tfmadapter}{https://github.com/AfrinDange/tfmadapter}.}

\begin{figure*}[t]
    \centering
    \includegraphics[width=0.95\textwidth]{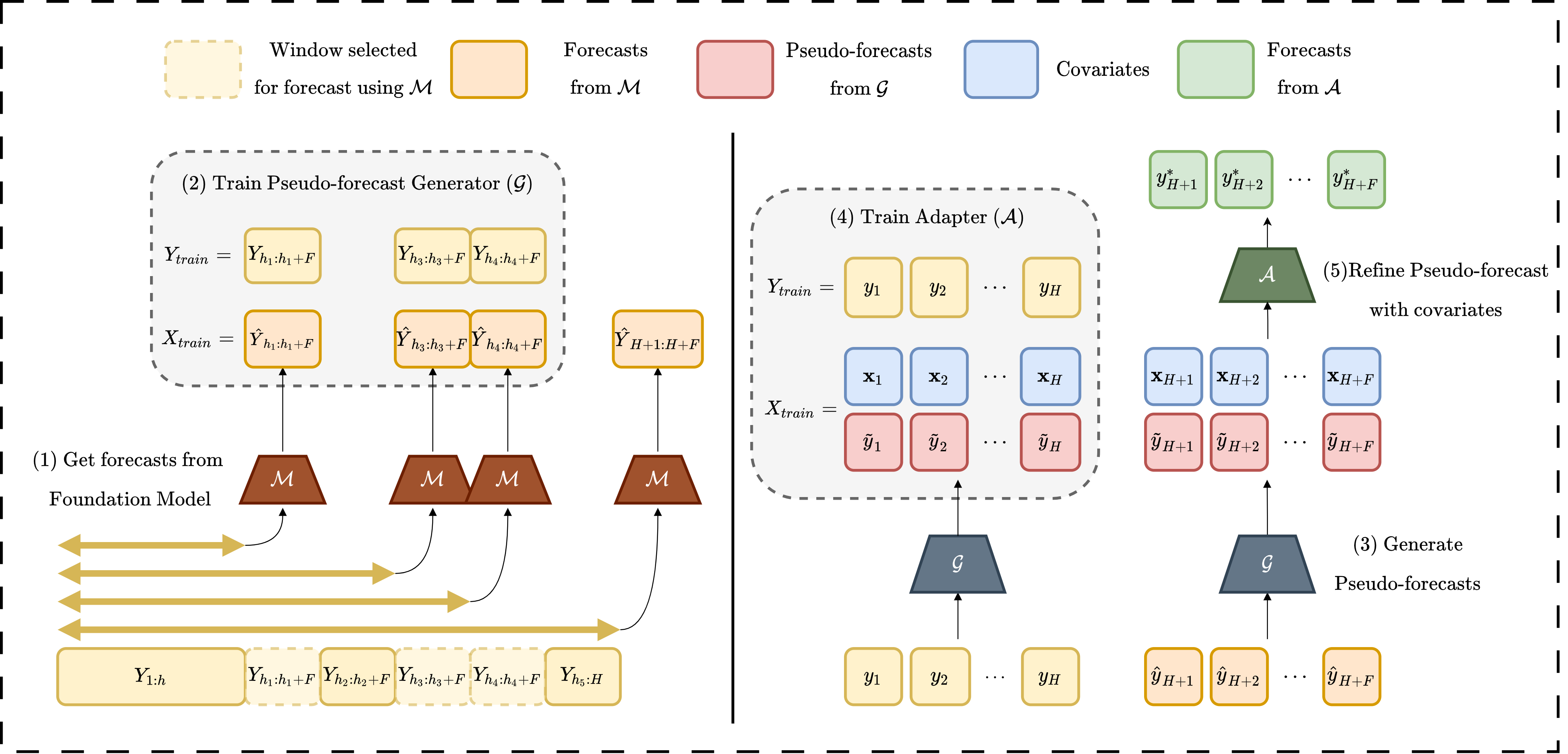}
    \caption{Overall \sysname{} Approach. $(1)$ Windows are selected based on the variance of true values to generate labeled data for training the pseudo-forecast Generator $\mathcal{G}$. $(2)$ The Foundation Model $\mathcal{M}$ produces forecasts over these windows, which, along with true values, serve as training data for the Generator. $(3)$ The trained Generator generates pseudo-forecasts for all time steps. $(4)$ The Adapter $\mathcal{A}$ is trained using pseudo-forecasts, covariates, and true values for time steps in history. $(5)$ The trained Adapter predicts the final forecast by integrating covariate signals with the Foundation Model’s predictions. Additional features are omitted for brevity.}
    \Description[Diagram of the \sysname{} Approach]{Overall \sysname{} Approach. $(1)$ Windows are selected based on the variance of true values to generate labeled data for training the pseudo-forecast Generator $\mathcal{G}$. $(2)$ The Foundation Model $\mathcal{M}$ produces forecasts over these windows, which, along with true values, serve as training data for the Generator. $(3)$ The trained Generator generates pseudo-forecasts for all time steps. $(4)$ The Adapter $\mathcal{A}$ is trained using pseudo-forecasts, covariates, and true values for time steps in history. $(5)$ The trained Adapter predicts the final forecast by integrating covariate signals with the Foundation Model’s predictions. Additional features are omitted for brevity.}
    \label{fig:adapter}
\end{figure*}

Time Series Foundation Models (TSFMs)~\cite{tsfm_survey} have recently been demonstrated to provide state-of-the-art accuracy in long-term forecasting for new time series just by conditioning on a history of past time series values as input context.  These models are 
%
trained on vast amounts of time series data from diverse domains, including energy, climate, web traffic, economics, finance, transportation, nature, and healthcare \cite{woo2024unified}. Due to their extensive training corpora, these foundation models acquire the appropriate inductive biases to generate quality forecasts with just a forward pass on the history of past time series values. These models even surpass custom supervised methods in univariate forecasting.

Time series forecasting can be categorized into three primary settings: univariate forecasting, multivariate forecasting, and forecasting with covariates. While most foundation models excel in univariate settings, only a few extend to multivariate forecasting, and even fewer integrate covariates. Covariates\textemdash exogenous variables such as weather conditions, economic indicators, and promotional events\textemdash are available for future time steps and often provide a critical context for improving forecast accuracy. Unlike historical time series values, covariates can be available in advance, allowing models to anticipate future variations that are not evident from past trends alone. Despite their significance, most existing foundation models do not support covariates, because covariates are highly domain-specific, and the covariates of one time series may be totally unrelated with those of another. 

Moirai \cite{woo2024unified} is one of the few foundation models incorporating covariates, but it does not adequately exploit the information in covariates. We show in Section ~\ref{sec:expt:fc} that even when the true label is provided as a covariate, the model fails to leverage the oracle signal to generate a perfect forecast. This reveals a fundamental limitation in current foundation models: their inability to effectively harness exogenous information. 

Our goal is to design an adapter that enhances the forecasting capabilities of foundation models by integrating covariates.  In order to retain the ease of usage of multi-tenancy foundation models, we require that the adapter be \emph{instance-level}, that is depend only on the historical data that is provided to the foundation model during one invocation. This makes our goal different from two recent models, ChronosX~\cite{arango2025chronosxadaptingpretrainedtime} and TTM~\cite{ekambaram2024tiny}, that require updating the parameters of the TSFM with domain-specific time series data that goes beyond the limited history provided to the TSFM during inference.
%
%
With instance-level learning, we seek to learn to cascade the covariates with the univariate forecasts of the TSFM, 
using the limited labeled data in the history.  Such learning poses significant challenges. First, labeled data is limited and will require univariate forecasts at each step in the historical data.  Second, foundation models are computationally expensive, requiring careful consideration for invoking them to generate such intermediate forecasts. Lastly, time series data is dynamic in nature, requiring the model to adapt to temporal variations without overfitting to specific patterns.


We present \sysname, a method of training lightweight Gaussian Processes to combine forecasts from the TSFM with covariates. Our main technical novelty lies in how we harness the entire history while making only a small number of calls to the TSFM.  We achieve this via a two-stage method.  In the first stage, we train a single-variable regression model to generate pseudo forecasts for the entire historical context. These pseudo-forecasts provide the complete feature set of the entire history by serving as a proxy for time steps where foundation model forecasts are not generated. In the second stage, we train a lightweight supervised regression model that integrates both the foundation model's forecast and the covariates, essentially to capture their relationship and refine the final predictions. We evaluate our approach on forecasting with covariates and demonstrate that our adaptation method achieves superior performance compared to supervised methods trained using only the historical context provided as input in the example, and foundation models without fine-tuning.

\paragraph{Contributions} Our contributions are as follows:
\begin{itemize}
	\item We identify and analyze the limitations of existing foundation models in harnessing covariates for time series forecasting.
	\item We propose an instance-level lightweight adaptation method that enables foundation models to harness covariates without fine-tuning them.
    \item We present a novel two-stage method of harnessing the entire history in an instance to train the adapter while making a small number of calls to the TSFM.
	\item We demonstrate the efficacy of our approach through experiments with several real datasets, showcasing significant improvements over several baseline methods.
\end{itemize}

\section{Related Work}
\subsection{Emergence of Time Series Foundation Models}
Time series forecasting has traditionally relied on statistical methods such as ARIMA \cite{box_pierce_arima} and ETS \cite{Hyndman2008ForecastingWE}. These methods offer interpretability and work well for stationary and seasonal data. However, they struggle with capturing complex dependencies and adapting to dynamic, high-dimensional datasets.

Deep learning introduced more expressive models for time series forecasting. Models like N-BEATS \cite{Oreshkin2020N-BEATS:}, N-HiTS \cite{Challu_nhits}, TCN \cite{chen_tcn}, and TiDE \cite{das2023longterm} leverage MLPs and convolutional networks to capture hierarchical and long-range dependencies. Recurrent architectures such as LSTM \cite{hochreiter1997_lstm} and GRU \cite{chung2014_gru} further enhanced forecasting by explicitly modeling sequential dependencies. Methods like DeepAR \cite{salinasdeepar} and LSTNet \cite{lai_lstnet} showcased the effectiveness of these recurrent models in handling time series data with complex temporal patterns. 

Following the emergence of Transformers \cite{vaswani2023attentionneed}, these models became the preferred choice for sequence modeling beyond NLP. Transformer-based forecasting models such as Informer \cite{haoyietal-informer-2021, haoyietal-informerEx-2023}, Autoformer \cite{wu2021autoformer}, and FEDformer \cite{zhou2022fedformer} were proposed to handle long-range dependencies. 
PatchTST~\cite{nie2023a} improved on these by introducing patching that divides time series into fixed-length patches of size $p$, reducing sequence length from $N$ to $\lceil N/p \rceil$. Attending to patches of time series helps capture temporal patterns better. More importantly, it demonstrated that Transformers trained on one dataset could generalize well to new datasets, showcasing their capability for transfer learning in time series forecasting. This insight led to foundation models for time series, which are pretrained on large-scale datasets and exhibit generalization to unseen time series.

\subsection{Transformer-based Foundation Models}
Foundation models based on transformers are built on the three architectural variations:

\textbf{Encoder-only:} Encoder-only foundation models leverage bidirectional attention by utilizing masked reconstruction for pretraining. Moirai \cite{woo2024unified} is pretrained by masking tokens primarily in the forecast horizon. It also handles multivariate forecasting by appending each series's context and masked tokens into one sequence. Multiple series and their time steps are distinguished using a variate ID and time ID. Forecasting with covariates follows a similar approach. Moment \cite{goswami2024moment} is a general-purpose time series foundation model pretrained with random masking of input tokens. Unlike Moirai, Moment heavily relies on fine-tuning or linear probing for forecasting and time series classification tasks. The reconstruction head pretrained alongside the encoder can be used directly for imputation and anomaly detection tasks.

\textbf{Decoder-only:} Decoder-only foundation models are typically pretrained using next-value prediction and perform inference autoregressively for sequences longer than the patch size. TimesFM \cite{das2024a}, a decoder-only foundation model, is pretrained using a fixed input patch size $(p_{\text{input}}=32)$ and a fixed output patch size $(p_{\text{output}}=128)$. Lag-Llama \cite{rasul2023lagllama} builds upon the LLaMA architecture \cite{touvron2023llamaopenefficientfoundation} and incorporates lag features during tokenization instead of utilizing patching. Incorporating lag features requires an extended history based on the number of lag features and does not reduce the sequence length.

\textbf{Encoder-Decoder: } TimeGPT \cite{garza2024timegpt1} is a proprietary foundation model based on the standard transformer architecture proposed by \cite{vaswani2023attentionneed}. Although it handles covariates, the specifics of the implementation are not provided. Chronos \cite{ansari2024chronos} utilizes T5 architecture \cite{raffelT5} and quantizes continuous time series values into discrete tokens using uniform binning. The latest version, \textit{Chronos-bolt}, incorporates patching at the input of the encoder and performs direct multi-step forecasting \footnote{\url{https://aws.amazon.com/blogs/machine-learning/fast-and-accurate-zero-shot-forecasting-with-chronos-bolt-and-autogluon/}}.

\subsection{Foundation Models Harnessing Covariates}
\begin{figure}[h]
    \centering
    \includegraphics[width=\linewidth]{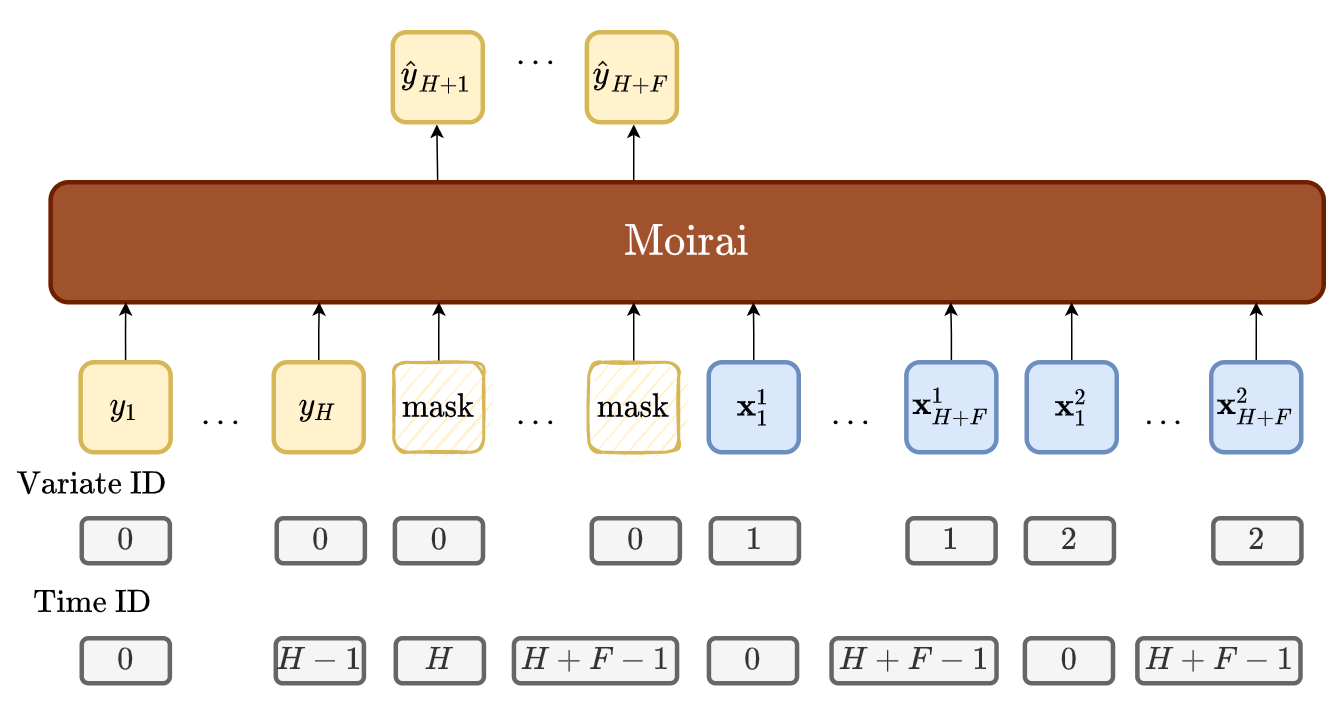}
    \caption{Input and output representation of Moirai for forecasting with covariates. For brevity, we assume patch size $p=1$. The time series values $\{y_1, y_2, \ldots, y_H\}$ are available as history along with covariates $\{x^i_1, x^i_2, \ldots, x^i_{H+1:H+F}\}$ for $i=\{1,2\}$.} 
    \Description[Input and output representation of Moirai for forecasting with covariates.]{Input and output representation of Moirai for forecasting with covariates. For brevity, we assume patch size $p=1$. The time series values $\{y_1, y_2, \ldots, y_H\}$ are available as history along with covariates $\{x^i_1, x^i_2, \ldots, x^i_{H+1:H+F}\}$ for $i=\{1,2\}$.}
    \label{fig:moirai}
\end{figure}

Tiny Time Mixers (TTM) \cite{ekambaram2024tiny} are foundation models based on MLPs. TTM offers fine-tuning a linear head for incorporating covariates. The only foundation model that incorporates covariates in its architectures is Moirai \cite{woo2024unified}. Figure~\ref{fig:moirai} illustrates how Moirai incorporates covariates for forecasting. ChronosX \cite{arango2025chronosxadaptingpretrainedtime} uses MLP blocks at the input and the output layer and requires fine-tuning to adapt covariates in foundation models\footnote{We have not included ChronosX in evaluation as their \href{https://github.com/amazon-science/chronos-forecasting/tree/ccb656cc2b67d69cef4a20c14183a634318f843b}{code} is not available yet.}. In contrast, our method is targeted for instance-level adaptation using only the data in the history of the current instance.


\section{Problem Definition}
We first define the time series forecasting problem and its common variants: univariate, multivariate, and forecasting with covariates. We then define our specific goal of instance-level adaptation of foundation models to handle covariates.

\subsection{Forecasting Settings}
Consider a set of $T$ time series, where $Y^t_{1:H}=\{y^t_1, y^t_2, \ldots, y^t_H\}$ denotes the sequence of $H$ observed values for the $t$-th time series. The goal of time series forecasting is to predict the future $F$ values, $\hat{Y}^t_{H+1:H+F} = \{\hat{y}^t_{H+1}, \hat{y}^t_{H+2}, \ldots, \hat{y}^t_{H+F}\}$. We categorize forecasting based on the input information.

\noindent\textbf{Univariate Forecasting:} This involves predicting the future values of a single time series using only its past observations:
$$\hat{Y}_{H+1:H+F} = f(Y_{1:H})$$

\noindent\textbf{Multivariate Forecasting:} When multiple $(T>1)$ time series are available up to time $H$, multivariate forecasting models predict the future values of a target series by considering the history of all available series: 
$$\hat{Y}^t_{H+1:H+F} = f(\{Y_{1:H}^i: i=1,2,\ldots, T\})$$

\noindent\textbf{Forecasting with Covariates:} This setting incorporates external variables, or covariates, denoted as $\vX_{1:H+F} = \{\vx_1, \vx_2, \dots, \vx_{H+F}\}$, which are available across both the history and the forecast horizon. If these covariates are informative and correlated with the target time series, they can significantly enhance forecasting performance. The prediction is then based on both the history of the target series and the past and future values of the covariates:
$$\hat{Y}_{H+1:H+F} = f(Y_{1:H}, \vX_{1:H+F})$$

\newcommand{\adapt}{\mathcal{A}}
\subsection{Instance-level Adapters for Forecasting with Covariates}
We assume the availability of a time series foundation model $\mathcal{M}$ that maps the history $Y_{1:H}$ directly to future values $\hat{Y}_{H+1:H+F} = \mathcal{M}(Y_{1:H})$ without requiring fine-tuning. 
However, most foundation models do not inherently incorporate covariates. Our objective is to enable foundation models to effectively leverage covariates while preserving their strong generalization capabilities. To improve upon the univariate forecasts, we need to integrate information from covariates $\vX_{1:H+F}$. For this we need to design an adapter $\adapt(\cM, Y_{1:H}, \vX_{1:H+F})$ that makes a limited number of calls to $\cM$, and leverages the history $Y_{1:H}$ only in the current instance to learn how to combine the univariate forecasts from the foundation model with the information in the covariates. 


The core challenge here is that generating sufficient, representative training data for this combiner model is difficult.  If we treat the combiner as a regression model, the history $Y_{1:H}$ provides the target output for training with inputs as the covariates and the forecasts from the TSFM $\cM$.  However, we cannot obtain the forecasts from $\cM$ on the entire history since $\mathcal{M}$ requires a certain minimum history of data in order to make meaningful forecasts on a new series.  Let $h$ be that minimum history. That leaves only $H-h$ instances for training the combiner, which may not suffice.  A second challenge is that $\cM$ makes forecasts for $F$ time-steps into the future. In order to get forecasts from $\cM$ in the $H-h$ span, we may require multiple expensive calls.  In the next section, we present our novel two-stage approach based on pseudo forecasts that harnesses the entire history for training the combiner, while making a constant number of calls to the $\cM$.


\section{Our Approach: \sysname}
We now present \sysname{}, a lightweight instance-level adaptation method designed to effectively harness covariates to augment the predictions of pretrained time series foundation models, while making only a constant number of additional calls to $\cM$ on subsets of the entire history. 
\sysname{} operates in two stages. First, we generate pseudo-forecasts to simulate the behavior of the foundation model across the entire history. Second, we use these pseudo-forecasts, along with covariates and auxiliary features, to train a regression-based adapter that refines the foundation model's outputs. The full procedure is outlined in Algorithm~\ref{alg:adapter}, and Figure~\ref{fig:adapter} presents an overview.

\begin{small}
\begin{algorithm}[htb]
    \caption{Lightweight Adapter for Forecasting with Covariates}
    \label{alg:adapter}
    \begin{algorithmic}
    \State \textbf{Input:} History \( Y_{1:H} \in \Re^{H \times 1} \), Covariates \( \vX_{1:H+F} \in \Re^{(H+F) \times d} \)
    \State \textbf{Output:} Forecast \( Y^*_{H+1:H+F} \in \Re^{F \times 1} \) \\

    \State \textbf{\textit{\large{Stage-I}}}
    \State \textbf{Acquire Labeled Data from the Foundation Model}
    \State Partition \({h:H} \) into \(N=(H-h)/F\) non-overlapping windows $W_i=(h_i+1,h_{i+F}):i=1\ldots N$, 
    \\ \Comment{h is the minimum history to predict reasonable forecasts.}
    \For{\(\{W_n\}_{n=1}^N \)}
        \State Compute \( Z_n = \frac{W_n - \mu_y}{\sigma_y} \) \\ \Comment{\(\mu_y\) and \(\sigma_y\) are mean and standard deviation of \(Y_{1:H}\)} 
        \State Compute \( \bar{Z}_n = \frac{1}{F} \sum_{i=1}^F Z_n(i) \)
    \EndFor
    \State Retain \(K=3\) windows with the lowest, median, and highest \( \bar{Z}_n \). \\

    \State \textbf{Generate Pseudo-forecasts}
        \For{each window \( (W_i)_{i=1}^{K} \) containing \( Y_{h_i+1:h_i+F} \)}
            \State Compute \( \hat{Y}_{h_i+1:h_i+F} = \cM(Y_{1:h_i}) \) \\ \Comment{$\cM$ is the foundation model}
        \EndFor
        
        \State Define feature set: 
        \(
            F = \{\text{lag-L}, \text{positions}\}
        \)
        where: \\
        \Comment{L is the number of lag features}
        \State \hspace*{\algorithmicindent}\(
            \text{lag-L}(y_t) = Y_{t-L:t-1} 
        \),
        \State \hspace*{\algorithmicindent}\(
            \text{positions}(t) = \text{sinusoidal}(t \mod \text{s}) \in \Re^p
        \) \\\Comment{s is seasonality}
        \State Construct training data: 
        \State \hspace*{\algorithmicindent}$
            \begin{aligned}
            \mathbf{X}_{\text{train}} = \big[& Y_{h_i+1:h_i+F};\ \text{lag-L}(Y_{h_i+1:h_i+F}); \\
                                             & \text{positions}(h_i+1:h_i+F) \big]_{i=1}^K,
            \end{aligned}$
        \State \hspace*{\algorithmicindent} \(
            Y_{\text{train}} = [\hat{Y}_{h_i+1:h_i+F}]_{i=1}^K
        \) 
        \State Train Pseudo-forecast Generator \(\mathcal{G}\) on \( (\mathbf{X}_{\text{train}}, Y_{\text{train}}) \)
        \State Compute \( \tilde{Y}_{1:H} = \mathcal{G}(Y_{1:H}) \)
        \State \phantom{Compute} \(\tilde{Y}_{H+1:H+F} = \mathcal{G}(\cM(Y_{1:H})) \) \\

    \State \textbf{\textit{\large{Stage-II}}}
    \State \textbf{Train the Adapter Model}
        \State Select optimal model \(\mathcal{A} \in \mathcal{GP}(m,k)\) and features \( f \subseteq F \) using validation on \(
            (\mathbf{X}_{\text{train}}=[\tilde{Y}_{1:H}; \vX_{1:H}; f], Y_{\text{train}}=Y_{1:H})
        \) 
    
        \State Train adapter model \( \mathcal{A} \) on 
        \[
            (X_{\text{train}}=[\tilde{Y}_{1:H}; \vX_{1:H}; f],\quad Y_{\text{train}}=Y_{1:H})
        \]
    
    \State \textbf{Compute Final Forecast}
        \State Compute forecast: 
        \[
            Y^*_{H+1:H+F} = \mathcal{A}(\tilde{Y}_{H+1:H+F}, \vX_{H+1:H+F}, f)
        \]
    \end{algorithmic}
\end{algorithm}
\end{small}

\subsection{Stage I: Acquiring Labeled Data}
The first stage addresses the challenge of limited training data. To overcome this, we first construct a proxy dataset that approximates the foundation model’s forecast behavior over the entire history. This allows us to train a regression model with features that mimic the foundation model’s predictions.

To efficiently obtain diverse samples of the foundation model's forecasts, we partition the available time series values $Y_{h:H}$\textemdash where $h$ is the minimum historical context required by $\cM$ to predict reasonable forecasts\textemdash into $N$ non-overlapping windows of length $F$, with each window $W_i = Y_{h_i+1:h_i+F}$ starting at $h_i+1$ such that $Y_{1:h_i}$ provides the context for forecasting with $\cM$. We normalize each window $W_n$ using the mean $\mu_y$ and standard deviation $\sigma_y$ of the entire history $Y_{1:H}$, yielding $Z_n = \frac{W_n - \mu_y}{\sigma_y}$. We then compute the average z-score for each window: $\bar{Z}_n = \frac{1}{F}\sum_{i=1}^{F} Z_n(i)$. This average z-score provides a measure of the window's overall deviation from the mean $\mu_y$. We then select $K=3$ representative windows, corresponding to the lowest, median, and highest average z-scores $\bar{Z}_n$, representing windows with low, medium, and high variability.

The foundation model $\cM$ is then invoked to produce forecasts for each selected window: $\hat{Y}_{h_i+1:h_i+F} = \cM(Y_{1:h_i})$. 

These forecasts serve as targets for the pseudo-forecast generator. For each time step $t$, we construct feature vectors comprising: (i) short-term lag features $Y_{t-L:t-1}$ where $L$ is the number of lag features, and (ii) sinusoidal positional encodings $\text{sinusoidal}(t \mod s)$, where $s$ denotes the seasonality of the time series.

We employ Bayesian Ridge regression as the pseudo-forecast generator $\mathcal{G}$, due to its ability to inherently capture uncertainty and impose prior distributions over parameters, making it ideal for limited training data. This generator is trained to predict forecast values for all time steps $t \in [1, H]$.

\subsection{Stage II: Training the Adapter}
In the second stage, we learn an adapter model that refines the foundation model's univariate forecasts by leveraging covariates. This stage aims to learn the conditional distribution of the true time series values given the foundation model's predictions and the available covariates.

We model the adapter using Gaussian Processes (GPs), which are non-parametric and provide a closed-form solution of the posterior distribution. Specifically, we train the adapter, $\mathcal{A} \in \mathcal{GP}(m,k)$ to learn the mapping:
$$y_t = g(\tilde{y}_t, \vx_t, f_t) + \epsilon_t, \quad \epsilon_t \sim \mathcal{N}(0, \sigma_t^2)$$

where $y_t$ is the true time series value, $\tilde{y}_t$ is the corresponding pseudo-forecast, $\vx_t$ are the covariates, and $f_t$ are the auxiliary features such as lag features and positional encodings. Here, $g$ is drawn from a GP prior and $\epsilon_t$ represents observation noise. 

We define a composite kernel $k(z,z') = k_1(z_1, z_1') + k_2(z_2,z_2')$. Here, $z = [z_1, z_2]$ where $z_1$ is the concatenation of the pseudo-forecasts, lag features, and positional encodings, and $z_2$ represents the covariates. $k_1$ models the relationship between the pseudo-forecast and the target variable, and $k_2$ models the direct influence of the covariates. We explore kernel choices such as Matern, RBF, and Linear, or a combination of these kernels.

We reserve the latest $F$ time steps, $H-F$ to $H$, for validation. We use this validation set to select the kernel and optimize its parameters. Once the optimal configuration is obtained, we train the adapter on the entire training dataset:
$$X_{\text{train}} = [\tilde{Y}_{1:H}; \vX_{1:H}; f], \quad Y_{\text{train}} = Y_{1:H}$$

The predictive distribution for the forecast horizon is given by the GP posterior in closed-form: 
$$\mu^* = K_*K^{-1}Y_{train},\quad \Sigma^*=K_{**} - K_*K^{-1}K_*^T,$$

\noindent where $K$ is the kernel matrix over training data, $K_*$ between training and test data, and $K_{**}$ over test data. The final forecast is:

$$Y^*_{H+1:H+F} = \mu^*$$

\noindent\textbf{Uncertainty-based Filtering:} By incorporating posterior predictive variance as an uncertainty estimate, we discard forecasts for which the variance $Var(y^*_t)$ exceeds a threshold (determined during validation) and revert to the foundation model's univariate forecast for that time step. This step allows the adapter to avoid using covariates when they are unreliable.

\section{Experiments}
In this section, we discuss our experimental settings and comprehensively evaluate our method on real datasets against foundation models and supervised baselines.

\subsection{Forecasting with Covariates Using Moirai}
\label{sec:expt:fc}
Moirai is a popular time series foundation model capable of operating in univariate, multivariate, and multivariate with covariates forecasting settings.  While its univariate forecasting performance has been extensively evaluated, we investigate its capabilities in multivariate forecasting and forecasting with covariates.


\subsubsection{Datasets} We conduct our experiments on four benchmark datasets: ETTh1, ETTh2, ETTm1, and ETTm2 \cite{haoyietal-informer-2021}. The datasets, comprising both hourly (ETTh1, ETTh2) and 15-minute (ETTm1, ETTm2) time series, have been used in prior work to demonstrate Moirai's effectiveness; it has also outperformed several supervised methods on them.

\begin{figure}[h]
    \centering
    \includegraphics[width=\linewidth]{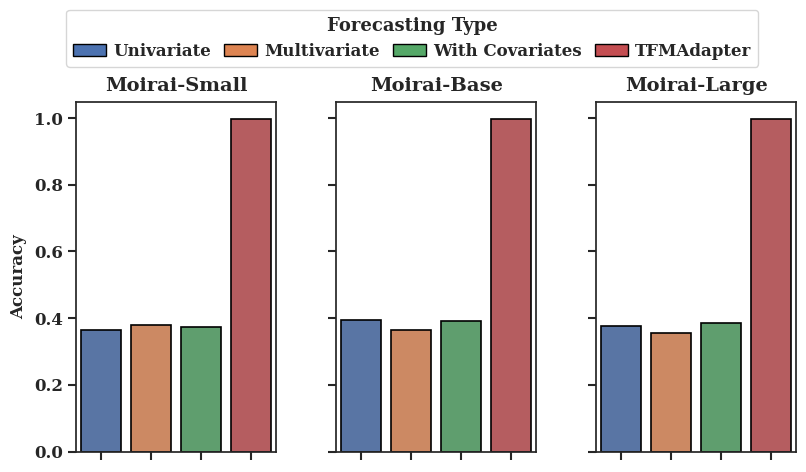}
    \caption{Average accuracy of Moirai foundation models on ETT datasets for univariate, multivariate, and covariate-enhanced forecasting.  The ideal accuracy ($y=1$) represents the expected performance if true labels were fully utilized. The gap between actual and ideal accuracy suggests Moirai does not fully leverage covariates, whereas \sysname{} achieves ideal accuracy.}
    \Description[Comparison of Moirai foundation model on ETT datasets]{Average accuracy of Moirai foundation models on ETT datasets for univariate, multivariate, and covariate-enhanced forecasting.  The ideal accuracy ($y=1$) represents the expected performance if true labels were fully utilized. The gap between actual and ideal accuracy suggests Moirai does not fully leverage covariates.}
    \label{fig:moirai_covariate_analysis}
\end{figure}

\subsubsection{Forecasting Settings} We evaluate three variants of the Moirai foundation model\textemdash small (13.8M parameters), base (91.4M parameters), and large (311M parameters)\textemdash across three settings: 
\begin{enumerate}
    \item \textbf{Univariate Forecasting:} Input: historical context $Y_{1:H}$ and output is forecasts $\hat{Y}_{H+1:H+F} = \cM(Y_{1:H})$.
    
    \item \textbf{Multivariate Forecasting:} Input: History of multiple time series $(Y^l_{1:H})_{l=1}^7$ and output is simultaneous forecasts of all  $(\hat{Y}^l_{H+1:H+F})_{l=1}^7 = \cM((Y^l_{1:H})_{l=1}^7)$.
    
    \item \textbf{Forecasting with Covariates:} To assess the model's ability to utilize covariates, we provide it with the history $Y_{1:H}$ along with covariates $\vX_{1:H+F}$, where we intentionally set $\vX_{1:H+F} = Y_{1:H+F}$. This approach provides the model with the true future values as covariates, allowing us to assess the potential performance gain when the oracle labels are available.
\end{enumerate}

Figure~\ref{fig:moirai_covariate_analysis} presents the evaluation accuracy for different Moirai model variants across these forecasting settings. Accuracy is calculated as 1 - SMAPE, averaged across the four ETT datasets. In this section, we report SMAPE by excluding time steps with zero true values. This is necessary in case of the ETT2 datasets, which contain many zeros. \sysname{} achieves near-zero MAE by accurately predicting these zero values when given the oracle label; however, SMAPE reflects an inflated error due to a zero denominator when both the true and the forecasted value are zero, despite the high accuracy.

Previous works \cite{liu2024itransformer, zhang2023crossformer} have shown that multivariate forecasting can outperform univariate forecasting on the ETT datasets. We observe this trend only in the small Moirai model. Furthermore, providing future values as covariates should, in principle, lead to near-perfect accuracy. However, while incorporating these covariates does reduce MAE, particularly for the larger models, Moirai's ability to effectively utilize this future information appears to be severely limited. 

\subsection{Evaluation on Covariates}

\subsubsection{Datasets}

\textbf{Electricity Price Forecasting (EPF) Datasets} The EPF datasets are a standard benchmark of five distinct datasets for evaluating forecasting methods that incorporate covariates. Each dataset represents a unique electricity market, containing day-ahead electricity prices as the target variable and two covariates (continuous time series variables), which are day-ahead values of two influential external variables like load forecast, wind, or solar generation forecast. These datasets span six years of hourly data. We follow the \texttt{train/validation/test} split used by \cite{OLIVARES2023884} for evaluation.

\noindent\textbf{ImputeBench Datasets} The ImputeBench datasets \cite{imputebench} are a popular benchmark for missing value imputation. Each dataset in the benchmark contains multiple inter-related time series. These datasets span multiple domains, including climate, hydrology, energy, sensor, and weather. To adapt these datasets for forecasting with covariates, for each target time series, we identify the $k$ most highly correlated time series within the same dataset to use as covariates. We carefully determine the history $H$ and forecast horizon $F$ for evaluation based on the seasonality and the overall size of the dataset.

\begin{table}[h]
    \centering
    \caption{Description of Datasets Used for Evaluation.}
    \label{tab:expt_setting}
    \resizebox{\linewidth}{!}{
    \begin{tabular}{c|ccc|cc}
        \toprule
        Dataset & \makecell{Test\\Windows} & \makecell{Covariates} & Freq & History & \makecell{Forecast\\Horizon} \\ 
         \midrule
        EPF - BE  & 728 & 2  &  1H & 672 & 24 \\ 
        EPF - DE  & 728 & 2  &  1H & 672 & 24 \\ 
        EPF - FR  & 363 & 2  &  1H & 672 & 24 \\ 
        EPF - NP  & 728 & 2  &  1H & 672 & 24 \\ 
        EPF - PJM & 728 & 2  &  1H & 672 & 24 \\ 
      Air Quality &  10 & 4  &  1H & 672 & 24 \\
        Bafu      &  30 & 4  & 30T & 960 & 48 \\
        Chlorine  &  10 & 10 &  5T & 288 & 24 \\
        Climate   &  30 & 4  &  1M & 672 & 24 \\
        Drift     &  10 & 4  &  6H & 448 & 28 \\
      Electricity &  40 & 4  & 15T & 960 & 96 \\
        Meteo     &  30 & 4  & 10T & 960 & 96 \\
            Temp  &  30 & 4  &  1D & 768 & 32 \\
        \bottomrule
    \end{tabular}
    }
\end{table}

\subsubsection{Experiment Setting} 
We experiment with the foundation models, the adapter, and the baselines using only the history, $Y_{1:H}$, and covariates $\vX_{1:H+F}$ available for each example. We have specified the length of history available to each model in Table~\ref{tab:expt_setting}, along with the forecast horizon, frequency of the dataset\footnote{1H = 1 hour, 30T = 30 minutes, 5T = 5 minutes, 1M = 1 month, 6H = 6 hours, 15T = 15 minutes, 10T = 10 minutes, 1D = 1 day.}, and the number of non-overlapping test windows created for evaluation. We train each baseline using the history $H$ (as specified in Table~\ref{tab:expt_setting}) on a forecasting setting of history=$h$ where $(h < H)$ (see Algorithm~\ref{alg:adapter}) and forecast horizon=$F$. We create training data for the history using rolling windows of size $F$ with stride $1$. Training is performed for a maximum of 1000 steps with a fixed batch size and a consistent random seed to ensure reproducibility. We keep the latest $F$ values from $H-F$ to $H$ as validation data for both supervised baselines and the adapter. We use Optuna to perform hyperparameter tuning. Foundation models are evaluated without fine-tuning. We use GIFT-Eval as the base evaluation framework for all experiments~\cite{aksu2024gifteval}.

\subsubsection{Evaluation Metrics}
We evaluate models using three standard metrics: Mean Absolute Error (MAE), Symmetric Mean Absolute Percentage Error (SMAPE), and Root Mean Squared Error (RMSE). Given the forecasted values $\{\hat{y}_{H+1}, \hat{y}_{H+2}, \ldots, \hat{y}_{H+F}\}$ and their corresponding ground truth values $\{y_{H+1}, y_{H+2}, \ldots, y_{H+F}\}$ for the forecast horizon $F$, these metrics are defined as follows:
\begin{align*}
\text{MAE} &= \frac{1}{F} \sum_{t=H+1}^{H+F} \left| y_t - \hat{y}_t \right| \\
\text{SMAPE} &= \frac{2}{F} \sum_{t=H+1}^{H+F} \frac{\left| y_t - \hat{y}_t \right|}{\left( |y_t| + |\hat{y}_t| \right)} \\
\text{RMSE} &= \sqrt{\frac{1}{F} \sum_{t=H+1}^{H+F} (y_t - \hat{y}_t)^2 }
\end{align*}

Each metric evaluates various aspects of forecast quality: MAE provides an intuitive measure of absolute errors, SMAPE offers a scale-invariant measure of error, and RMSE penalizes larger errors more heavily. SMAPE is mainly used in ablation studies to compare relative improvements across different settings. 

\subsubsection{Baseline}
The baseline methods include established supervised models that are trained directly on the forecasting task without leveraging pretrained foundation models.


\begin{table*}
\caption{Performance comparison of \sysname, foundation models, and supervised baselines across multiple datasets. The best results are highlighted in \colorbox{green!20}{green}, and the second-best in \colorbox{yellow!20}{yellow}. \sysname{} consistently improves the performance of foundation models by effectively incorporating covariates. Moirai does not improve when forecasting with covariates, suggesting it fails to utilize them effectively. Supervised models, TimeXer and NBEATSx, do not perform well due to limited data.}
\label{tab:epf-main-results}
\resizebox{\textwidth}{!}{
\begin{tabular}{cc|rr|rrr|rr|rrrr}
\toprule
 &   & \multicolumn{7}{c|}{Foundation Models} & \multicolumn{4}{c}{Baseline} \\
 \midrule
\multirow[c]{2}{*}{Dataset} & \multirow[c]{2}{*}{Metric} & \makecell{Chronos\\\small{Adapter}} & \makecell{Chronos\\\small{Univariate}} & \makecell{Moirai\\\small{Adapter}} & \makecell{Moirai\\\small{Covariate}} & \makecell{Moirai\\\small{Univariate}} & \makecell{TimesFM\\\small{Adapter}} & \makecell{TimesFM\\\small{Univariate}} & \makecell{KReg} & \makecell{TiDE} & \makecell{NBE-\\ATSx-I} & \makecell{TimeXer}\\
\midrule
\multirow[c]{2}{*}{Air Quality} & MAE & \cellcolor{green!20}0.217 & 0.756 & \cellcolor{yellow!20}0.221 & 0.628 & 0.675 & 0.222 & 0.565 & 0.252 & 0.591 & 0.583 & 0.691 \\
 & RMSE & \cellcolor{yellow!20}0.680 & 1.230 & 0.680 & 1.033 & 1.099 & \cellcolor{green!20}0.676 & 0.977 & 0.722 & 0.941 & 1.029 & 1.023\\
\cline{1-13}
\multirow[c]{2}{*}{Bafu} & MAE & 0.135 & 0.140 & 0.136 & 0.145 & 0.135 & \cellcolor{green!20}0.131 & \cellcolor{yellow!20}0.133 & 0.182 & 0.238 & 0.153 & 0.253 \\
 & RMSE & \cellcolor{green!20}0.264 & 0.280 & 0.273 & 0.281 & 0.278 & \cellcolor{yellow!20}0.264 & 0.276 & 0.336 & 0.391 & 0.292 & 0.443 \\
\cline{1-13}
\multirow[c]{2}{*}{Chlorine} & MAE & 0.012 & 0.029 & \cellcolor{yellow!20}0.011 & 0.028 & 0.026 & \cellcolor{green!20}0.010 & 0.018 & 0.019 & 0.048 & 0.025 & 0.040 \\
 & RMSE & 0.025 & 0.051 & \cellcolor{green!20}0.022 & 0.053 & 0.048 & \cellcolor{yellow!20}0.022 & 0.033 & 0.038 & 0.070 & 0.057 & 0.057 \\
\cline{1-13}
\multirow[c]{2}{*}{Climate} & MAE & \cellcolor{green!20}0.322 & 0.406 & 0.351 & 0.475 & 0.474 & \cellcolor{yellow!20}0.348 & 0.474 & 0.348 & 0.381 & 0.350 & 0.455 \\
 & RMSE & \cellcolor{yellow!20}0.498 & 0.634 & 0.528 & 0.692 & 0.698 & 0.533 & 0.709 & \cellcolor{green!20}0.489 & 0.592 & 0.539 & 0.675 \\
\cline{1-13}
\multirow[c]{2}{*}{Drift} & MAE & \cellcolor{green!20}0.043 & 0.601 & \cellcolor{yellow!20}0.043 & 0.725 & 0.738 & 0.049 & 0.352 & 0.102 & 0.344 & 0.353 & 0.474 \\
 & RMSE & \cellcolor{yellow!20}0.087 & 0.814 & \cellcolor{green!20}0.085 & 0.953 & 0.988 & 0.102 & 0.538 & 0.233 & 0.523 & 0.494 & 0.695 \\
\cline{1-13}
\multirow[c]{2}{*}{EP-BE} & MAE & \cellcolor{green!20}6.376 & \cellcolor{yellow!20}6.430 & 6.586 & 6.890 & 6.732 & 6.538 & 6.587 & 8.921 & 10.373 & 9.498 & 10.565 \\
 & RMSE & 16.565 & \cellcolor{yellow!20}16.531 & \cellcolor{green!20}16.139 & 16.968 & 16.615 & 16.883 & 16.639 & 20.537 & 23.034 & 19.881 & 23.282 \\
\cline{1-13}
\multirow[c]{2}{*}{EP-DE} & MAE & 4.328 & 5.397 & 4.382 & 5.719 & 5.652 & \cellcolor{yellow!20}4.285 & 5.526 & \cellcolor{green!20}4.226 & 9.264 & 6.224 & 8.608 \\
 & RMSE & \cellcolor{green!20}7.421 & 8.976 & 7.653 & 9.626 & 9.454 & \cellcolor{yellow!20}7.423 & 9.391 & 7.498 & 14.199 & 10.107 & 13.385 \\
\cline{1-13}
\multirow[c]{2}{*}{EP-FR} & MAE & 4.610 & \cellcolor{green!20}4.497 & 4.976 & 4.757 & 4.815 & 4.679 & \cellcolor{yellow!20}4.547 & 7.810 & 7.764 & 5.343 & 7.806 \\
 & RMSE & \cellcolor{yellow!20}15.556 & 16.392 & 17.141 & 16.150 & 16.298 & \cellcolor{green!20}15.090 & 16.162 & 21.462 & 19.249 & 16.134 & 18.905\\
\cline{1-13}
\multirow[c]{2}{*}{EP-NP} & MAE & \cellcolor{green!20}1.916 & 2.027 & 2.010 & 2.156 & 2.140 & \cellcolor{yellow!20}1.990 & 2.114 & 2.520 & 3.664 & 2.686 & 3.485 \\
 & RMSE & \cellcolor{green!20}3.632 & \cellcolor{yellow!20}3.816 & 3.833 & 3.988 & 4.000 & 3.822 & 4.013 & 4.308 & 6.149 & 4.865 & 5.923 \\
\cline{1-13}
\multirow[c]{2}{*}{EP-PJM} & MAE & \cellcolor{green!20}3.394 & 3.621 & \cellcolor{yellow!20}3.494 & 3.869 & 3.747 & 3.499 & 3.714 & 4.025 & 5.447 & 4.207 & 5.515 \\
 & RMSE & \cellcolor{green!20}5.561 & 5.948 & \cellcolor{yellow!20}5.771 & 6.394 & 6.176 & 5.794 & 6.040 & 6.403 & 8.703 & 6.772 & 8.888 \\
\cline{1-13}
\multirow[c]{2}{*}{Electricity} & MAE & 0.276 & 0.271 & 0.293 & 0.305 & 0.301 & \cellcolor{yellow!20}0.228 & \cellcolor{green!20}0.218 & 0.329 & 0.381 & 0.390 & 0.412 \\
 & RMSE & 0.466 & 0.458 & 0.482 & 0.489 & 0.482 & \cellcolor{yellow!20}0.372 & \cellcolor{green!20}0.351 & 0.515 & 0.600 & 0.618 & 0.647 \\
\cline{1-13}
\multirow[c]{2}{*}{Meteo} & MAE & 0.280 & 0.348 & 0.282 & 0.394 & 0.355 & \cellcolor{yellow!20}0.277 & 0.342 & \cellcolor{green!20}0.274 & 0.504 & 0.431 & 0.587 \\
 & RMSE & 0.463 & 0.554 & 0.456 & 0.593 & 0.547 & \cellcolor{yellow!20}0.451 & 0.540 & \cellcolor{green!20}0.445 & 0.740 & 0.679 & 0.883 \\
\cline{1-13}
\multirow[c]{2}{*}{Temp} & MAE & \cellcolor{yellow!20}0.082 & 0.196 & 0.084 & 0.360 & 0.209 & 0.086 & 0.222 & \cellcolor{green!20}0.078 & 0.355 & 0.358 & 0.274 \\
 & RMSE & \cellcolor{yellow!20}0.115 & 0.252 & 0.117 & 0.469 & 0.268 & 0.122 & 0.285 & \cellcolor{green!20}0.107 & 0.482 & 0.726 & 0.356 \\
\cline{1-13}
\rowcolor{gray!10}
\multicolumn{2}{c}{\makecell{Average Performance\\Gain (in MAE)}} & \multicolumn{2}{|l}{27.267\%} & \multicolumn{3}{|l}{27.891\%} & \multicolumn{2}{|l}{24.961\%} & \multicolumn{4}{|l}{} \\
\bottomrule
\end{tabular}
}
\end{table*}

\textbf{Regression}  We train a kernel regression model with an RBF kernel and L2 regularization strength of $\alpha=1$, $f_k$ predicts the target value $y_t = f_k(\vx_i,\text{lag-L}(y_t))$ autoregressively, using the covariates and lag features.

\textbf{TimeXer} \cite{wang2024timexer} is a transformer-based supervised forecasting model that supports covariates. It performs dual attention where the dependencies within the time series are captured by self-attention, and covariates are integrated using cross-attention. We use a single encoder layer with two attention heads and a feed-forward layer with a hidden dimension of $64$. The model is trained with a patch size of $16$ and a hidden size selected from $64, 128$. We also optimize the dropout probability and learning rate.

\textbf{NBEATSx} \cite{OLIVARES2023884} extends the NBEATS \cite{Oreshkin2020N-BEATS:} model by harnessing covariates. We train NBEATSx-I containing identity, seasonal, and trend stacks. This is an interpretable version of the model, whereas NBEATSx-G is a general version that uses only identity stacks. Each stack contains a single block and uses a two-layer MLP with a hidden size of $128$. The model is trained with ReLU activations and a fixed hidden state size of $128$. We optimize dropout probability and learning rate.

\textbf{TiDE} \cite{das2023longterm} is an MLP-based supervised forecasting model that incorporates covariates. Given limited training data, we utilize a small model with a single encoder and decoder MLP layer. Additionally, we tune the hidden state size of the MLP, the learning rate, and the dropout probability.

\subsubsection{Foundation Models}  
We evaluate the following foundation models based on their strong performance across benchmarks and accessibility for experimentation.

\textbf{TimesFM} \cite{das2024a} We use the latest public version $2.0$ of TimesFM, supporting a context length of up to $2048$ time steps and comprises 500M parameters. Point forecasts from the model are used to generate labeled examples for the pseudo-forecast generator.  

\textbf{Chronos} \cite{ansari2024chronos} We experiment with the latest version, specifically the \textit{Chronos-Bolt-Base} model, which contains 205M parameters. This version incorporates patching and performs direct multi-step forecasting, improving efficiency for large histories and extended forecast horizons. We use the median of sampled predictions as labeled data.

\textbf{Moirai} \cite{woo2024unified} We use version 1.1 of the Moirai base model, which has 91.4M parameters. Similar to Chronos, we utilize the median of sampled predictions as labeled data. Additionally, we evaluate Moirai in a forecasting-with-covariates setting, using the same covariates provided to the adapter method.  

\subsubsection{\sysname{}}
In all experiments, \sysname{} uses a Bayesian Ridge regression model as the pseudo-forecast generator. It is trained to predict the foundation model's forecasts using lag features, sinusoidal position encodings, and the true time series values. The Adapter model is a Gaussian Process with kernels chosen from Matern, RBF, and Linear. The length scale and variance parameter of the kernels are optimized using validation. We use the closed-form posterior distribution for inference and use the obtained variance to filter highly uncertain forecasts. The threshold is tuned during validation.

\subsubsection{Results}
Table~\ref{tab:epf-main-results} reports the forecasting performance across several datasets, evaluated through MAE and RMSE. The results clearly indicate that foundation models equipped with \sysname{} generally outperform both the base foundation models and the supervised baselines across a majority of the datasets. For instance, Chronos and TimesFM, when augmented with the adapter to leverage covariates, achieve remarkable performance improvement on datasets such as Air Quality, Climate, Drift, EPF-DE, and Temperature. This highlights the efficacy of the adapter in effectively integrating covariates.

Moirai does not exhibit a consistent and significant performance gain when forecasting with covariates. In some instances, forecasting with covariates even leads to degradation in performance compared to univariate forecasting. For example, on the Bafu, EP-DE, EP-PJM, Meteo, and Temperature datasets, Moirai's performance with covariates is worse despite the adapter demonstrating an improvement over the univariate forecasts. 

The supervised baselines, TimeXer, NBEATSx, and TiDE, generally underperform compared to the foundation models. This is likely due to the limited amount of training data available to these models, highlighting the advantage of using pretrained foundation models. The Kernel Regression baseline performs notably well on the EPF-DE dataset.

\paragraph{Indirect Comparison with ChronosX~\cite{arango2025chronosxadaptingpretrainedtime}}
ChronosX is the latest foundation model that supports covariates with fine-tuning of MLP adapter layers. Since their code is not released, we present an indirect comparison by comparing our method with NBEATSx-G on the MASE metric that they used in \cite{arango2025chronosxadaptingpretrainedtime}.

\begin{table}[hbt]
    \centering
    \caption{Comparison of MASE for \sysname{}, ChronosX and NBEATSx-G. Dataset-level adaptation results are from \cite{arango2025chronosxadaptingpretrainedtime}}
    \label{tab:chronosx_comparison}
    \begin{tabular}{rr|rr}
        \toprule
         \multicolumn{2}{c|}{Instance-level Adaptation} & \multicolumn{2}{c}{Dataset-level Adaptation} \\
         \midrule
         \makecell{Chronos\\Adapter} & NBEATSx-G & ChronosX & NBEATSx-G \\
         \midrule
         0.666 & 0.891 & 0.420 & 0.412 \\
         \bottomrule
    \end{tabular}
\end{table}

NBEATSx-G outperforms ChronosX when fine-tuned at the dataset level, as shown in Table~\ref{tab:chronosx_comparison}. However, since our method uses instance-level data, a direct comparison with ChronosX under dataset-level adaptation is not meaningful. We evaluate NBEATSx-G under instance-level adaptation. \sysname{} achieves better performance than instance-level NBEATSx-G, suggesting \sysname{} would likely outperform ChronosX if evaluated at the instance level.

\section{Ablation Studies}
To systematically evaluate the impact of the design choices in \sysname{}, we conduct ablation studies focusing on two key components: (1) the window selection strategy for invoking foundation models, and (2) the effectiveness of the pseudo-forecast generator in capturing the relationship between the true values and the forecasts generated by the foundation models.

\subsection{Pseudo-forecasts}
To isolate the contribution of the pseudo-forecast generator, we perform an ablation where it is removed from the architecture. Instead, the Adapter is trained directly using forecasts obtained from the foundation model. We further examine the impact of varying the amount of training data, which corresponds to increasing the number of foundation model invocations. We experiment with $k={3,5,8}$ foundation model invocations, which result in training data of size $kF$. SMAPE is used as the evaluation metric, and the reported results are averaged over all EPF and ImputeBench datasets.

\begin{figure}[h]
    \centering
    \includegraphics[width=\linewidth]{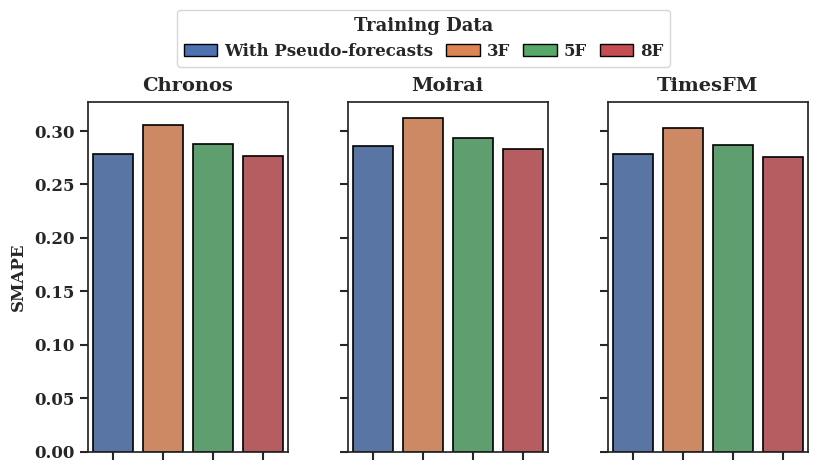}
    \caption{Impact of the Pseudo-Forecast Generator. Training with pseudo-forecasts achieves performance comparable to directly training the Adapter with forecasts obtained from the foundation model using $8F$ training data, and outperforms training with $3F$ and $5F$ training data.}
    \Description[Impact of the Pseudo-Forecast Generator]{Impact of the Pseudo-Forecast Generator. Training with pseudo-forecasts achieves performance comparable to directly training the Adapter with forecasts obtained from the foundation model using $8F$ training data, and outperforms training with $3F$ and $5F$ training data.}
    \label{fig:ablation:pseudo_forecasts}
\end{figure}

Figure~\ref{fig:ablation:pseudo_forecasts} demonstrates that training the adapter using pseudo-forecasts achieves performance comparable to directly training the adapter on the foundation model's forecasts obtained via $k=8$ invocations. When the adapter is trained directly on a limited amount of foundation model forecasts ($k={3,5}$), the performance is worse compared to using the pseudo-forecast generator, which itself uses $k=3$ foundation model invocations to acquire labeled data for training. As it is a regression model with a closed-form solution, the pseudo-forecast generator offers a highly efficient means of leveraging the entire history and enabling the adapter to attain a level of performance comparable to using a significant number of computationally intensive foundation model invocations. 
\subsection{Window Selection}
The main purpose of the window selection mechanism is to provide representative samples within the history $H$ that accurately capture the foundation model's behavior. These samples are used as training data for the pseudo-forecast generator; thus, the accuracy of the pseudo-forecast generator heavily relies on the selected samples. We evaluate three window selection strategies: 

Given the forecasting setting where history is $H$, forecast horizon is $F$, and the minimum context required to forecast accurately is $h$, we partition the interval $[H-h, H]$ into non-overlapping windows of size $F$.

\begin{enumerate}[leftmargin=*, align=left]
    \item \textbf{Latest} selects the $k=3$ most recent windows, specifically the windows spanning from time steps $H-kF$ to $H$.
    \item \textbf{Random} selects $k=3$ windows uniformly at random.
    \item \textbf{Z-score} selects three windows with the lowest, median, and highest average z-scores.
\end{enumerate}

\begin{figure}[h]
    \centering
    \includegraphics[width=\linewidth]{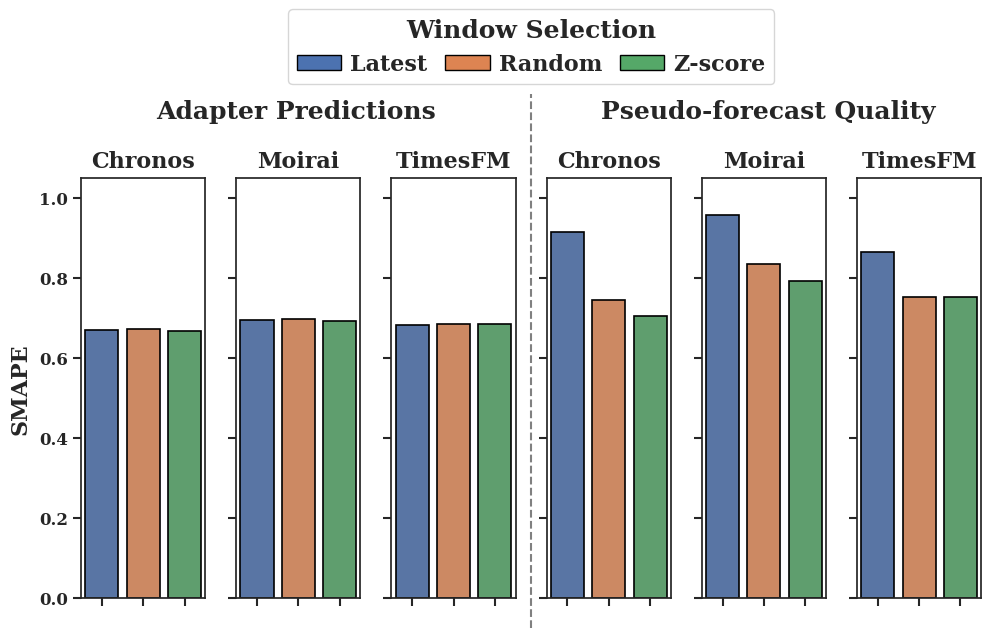}
    \caption{Comparison of Window Selection Strategies. The final forecast accuracy is similar across strategies, with the Latest strategy slightly outperforming others. The Z-score strategy, however, yields higher-quality pseudo-forecasts.}
    \Description[Comparison of Window Selection Strategies.]{Comparison of Window Selection Strategies. The final forecast accuracy is similar across strategies, with the Latest  strategy slightly outperforming others. The Z-score strategy, however, yields higher-quality pseudo-forecasts.}
    \label{fig:ablation-window-selection}
\end{figure}

We evaluate these strategies by comparing both the final forecasts and the pseudo-forecasts. Final forecasts are evaluated against the true values for the forecast horizon. Pseudo-forecast quality is evaluated against the true values for the history. We use SMAPE as the evaluation metric and average the results over all EPF and ImputeBench datasets.

As depicted in Figure~\ref{fig:ablation-window-selection}, the forecasting with covariates performance across Chronos, Moirai, and TimesFM remains relatively consistent regardless of the window selection strategy for acquiring training data for the pseudo-forecast generator. The latest strategy shows a marginal advantage. However, the quality of pseudo-forecasts, evaluated against the true values, varies significantly. The z-score strategy produces the most accurate pseudo-forecasts, but does not show significant improvement in the quality of the final forecasts. This suggests that producing pseudo-forecasts that more closely resemble the true data distribution does not necessarily improve the final forecasts, as the pseudo-forecasts may diverge from the predictive distribution of the foundation model.
\subsection{Varying History \& Forecast Horizon}
We further investigate the robustness of the Adapter by increasing both the history $H$ and the forecast horizon $F$. This analysis helps to elucidate \sysname{}'s performance in forecasting settings with longer historical contexts and forecast horizons. MAE is the reported evaluation metric, and the results are averaged across all EPF datasets.

\begin{table}[h]
\caption{Comparison of \sysname{} and Foundation Models for varying history (H) and forecast horizon (F). The ``Gain'' indicates the \% improvement in MAE achieved by the adapter.}
\label{tab:varying_hf}
\begin{tabular}{llrrrr}
\toprule
Model    & $H$   & $F$   & Adapter & Univariate & Gain \\
\midrule
\multirow{5}{*}{Chronos}
    & \multirow{3}{*}{672}  & 24  & 4.14 & 4.39 & 5.67\% \\
    &                        & 96  & 4.75 & 5.32 & 10.50\% \\
    &                        & 168 & 5.11 & 5.83 & 11.84\% \\
    \cmidrule(lr){2-6}
    & 1344                   & 24  & 4.13 & 4.37 & 5.02\% \\
    \cmidrule(lr){2-6}
    & 2016                   & 24  & 4.22 & 4.35 & 2.79\% \\
\midrule
\multirow{5}{*}{Moirai}
    & \multirow{3}{*}{672}  & 24  & 4.26 & 4.62 & 7.38\% \\
    &                        & 96  & 4.91 & 5.42 & 9.34\% \\
    &                        & 168 & 5.16 & 6.00 & 13.13\% \\
    \cmidrule(lr){2-6}
    & 1344                   & 24  & 4.30 & 4.54 & 5.06\% \\
    \cmidrule(lr){2-6}
    & 2016                   & 24  & 4.34 & 4.52 & 3.90\% \\
\midrule
\multirow{5}{*}{TimesFM}
    & \multirow{3}{*}{672}  & 24  & 4.17 & 4.50 & 7.17\% \\
    &                        & 96  & 4.88 & 5.42 & 10.04\% \\
    &                        & 168 & 5.20 & 5.93 & 11.96\% \\
    \cmidrule(lr){2-6}
    & 1344                   & 24  & 4.17 & 4.40 & 4.73\% \\
    \cmidrule(lr){2-6}
    & 2016                   & 24  & 4.23 & 4.38 & 3.40\% \\
\bottomrule
\end{tabular}
\end{table}

The results in Table~\ref{tab:varying_hf} consistently demonstrate a performance gain for the adapter compared to the base foundation models across different historical contexts and forecast horizons. Although the magnitude of the gain varies for different configurations of $H$ and $F$, the adapter consistently provides an improvement in forecasting accuracy. This indicates that the adapter effectively leverages covariates across varying history and forecast horizon lengths.

\section{Conclusion}
In this work, we examined the limitations of existing time series foundation models in leveraging covariates for forecasting. Our analysis revealed that even models such as Moirai, which are designed to support covariates, fail to effectively utilize them\textemdash even when oracle labels are provided. To address this, we proposed \sysname{}, a lightweight two-stage adaptation method that augments foundation models without modifying their parameters.

\sysname{} first generates pseudo-forecasts using a regression model trained on representative windows sampled from the history, where the labels for these windows are obtained by invoking the foundation model. These pseudo-forecasts are then combined with covariates and auxiliary features through a Gaussian Process-based Adapter to produce the final forecast. Our approach improves upon the base foundation models across a range of real-world datasets, outperforming both the foundation models themselves and supervised baselines while incurring minimal computational overhead.

These findings highlight the potential of adaptation techniques in bridging the gap between foundation and task-specific models. Future work could explore new methods for adapter models.

\section*{Acknowledgement}
We acknowledge the support of the SBI Foundation Hub for Data Science \& Analytics, and the Centre for Machine Intelligence and Data Science (C-MInDS) at the Indian Institute of Technology Bombay for providing financial support and infrastructure for conducting the research presented in this paper.

\section*{GenAI Usage Disclosure}
Portions of this writing were refined using generative AI tools to enhance clarity and readability. Specifically, OpenAI’s ChatGPT, Google’s Gemini 2.5 Flash, and Grammarly’s AI writing assistant were used for editing. The authors retain full responsibility for the accuracy and integrity of all content presented in this paper.

\bibliographystyle{ACM-Reference-Format}
\balance




\end{document}